# Detection of irregular QRS complexes using Hermite Transform and Support Vector Machine

Zoja Vulaj, Miloš Brajović, Anđela Draganić, Irena Orović

University of Montenegro, Faculty of Electrical Engineering

Džordža Vašingtona bb, 81000 Podgorica, Montenegro

email: zojavulaj@gmail.com

*Abstract* - Computer based recognition and detection of abnormalities in ECG signals is proposed. For this purpose, the Support Vector Machines (SVM) are combined with the advantages of Hermite transform representation. SVM represent a special type of classification techniques commonly used in medical applications. Automatic classification of ECG could make the work of cardiologic departments faster and more efficient. It would also reduce the number of false diagnosis and, as a result, save lives. The working principle of the SVM is based on translating the data into a high dimensional feature space and separating it using a linear classificator. In order to provide an optimal representation for SVM application, the Hermite transform domain is used. This domain is proved to be suitable because of the similarity of the QRS complex with Hermite basis functions. The maximal signal information is obtained using a small set of features that are used for detection of irregular QRS complexes. The aim of the paper is to show that these features can be employed for automatic ECG signal analysis.

## I. INTRODUCTION

The electrocardiogram (ECG) represents the measure of the electrical activity of the heart over certain period time. The ECG signals are periodic in nature because they are characterized by a sequence of periodical waves. The most important part of ECG signals are the QRS complexes [1]. The phenomenon related to QRS complex is the depolarization of the ventricles. Based on the analysis of QRS complexes, different health diagnoses can be established.

The analysis of recorded ECG data, especially in the case of patient monitoring, when a huge amount of ECG data is acquired, can be time-demanding. One such a case happens with patients affected by obstructive sleep apnea, which is the phenomenon of cessation of breathing during the sleep [2], [3]. The main reason of diseased people not being diagnosed with sleep apnea is because a cardiologist would be required to monitor the patient overnight. Because of this, the fast response to the demanding tasks and efficient analysis of ECG data becomes a challenge. For the purpose of making the work of cardiologist easier and decreasing the rate of undiagnosed (or improperly diagnosed) irregularities, computer based analysis of ECG signals are developed.

Artificial Neural Networks (ANN) proved to be a useful tool for solving classification tasks. The SVM have shown good performances, as well. Their application includes a large variety of cases: voice recognition, verification of signatures, diagnosis in medicine and microbiology, and even in automatic translation machines [4]. In this paper, the ECG signal processing using Support Vector Machines (SVM) is presented. The detection of abnormal QRS complexes is based on the recognition of two different data sets by comparing the features that are used to represent the data [4]. The basic features included into most classifications are the R peaks. When the R peaks are not sufficient for obtaining the desired results, new or additional features need to be used. Features can be divided into three types: temporal, morphological and statistical. Temporal features include the interval between the beats or the waves in ECG signals, as well as the heart rate. The heart rate is the key information for detecting arrhythmia. Morphological features characterize the morphology of the signal parts and are usually represented using coefficients obtained when the signal is decomposed into some transformation domain. Statistical features represent the statistical data of the signal (mean, maximum, minimum, etc). If all the features were used, their number would be so large that it would result in heavy computation. The difficulty faced during feature selection is to determine the right ones and the best combination of them. The features must be chosen such that the maximal signal information can be obtained from a small set of them. Consequently, different domains are used to achieve an optimal compact signal representation: the time domain, frequency domain, their combination, or some other domain [6]-[9]. Due to the similarity of the shapes of ECG signals and Hermite basis functions, in this paper QRS complexes are represented in the Hermite transform domain [10]. This transform domain is very suitable because the signal is presented using a small amount of coefficients while retaining the signal information. This fact indicates that the Hermite transform can be appropriate for ECG signal analysis, for either classification or abnormality detection. The selected features later represent the input of the classifier.

The paper is organized as follows: In Section II the theoretical background on the Hermite transform (HT) is given. The theory behind the SVM is explained in Section III. In section IV the experimental results are presented. The concluding remarks can be found in Section V.

## II. THE HERMITE TRANSFORM

ECG signals can be approximated with a reduced number of coefficients, while keeping the signal information. For this purpose, the Hermite transform domain is extensively exploited [6], [11], [12]. Consider an ECG signal $ECG(t)$. Since we use QRS complexes, our target signal of interest can be defined as

$QRS(t)$, where $|t| < T/2$ and $T$ is the definition interval of the QRS signal. The Hermite functions can be expressed using Hermite polynomials. The $n$-th order polynomial is defined using equation (1):

$$P_n(t) = (-1)^n e^{t^2} \frac{d^n(e^{-t^2})}{dt^n}. \quad (1)$$

The Hermite functions can be expressed as follows:

$$H_n(t) = \frac{e^{-t^2} P(t)}{\sqrt{2^n n! \sqrt{\pi}}}. \quad (2)$$

Since ECG signals (and in particular the QRS complexes) are continuous in time, in order to obtain an error-free approximation, the number of Hermite functions used in the signal expansion must be infinite. However, sampling both basis functions and signal at points $t_z$, $z = 0, \ldots, M-1$ proportional to the roots of the Hermite polynomial (1), the Hermite expansion becomes a finite discrete transform of the analyzed signal. In this case each QRS complex can be uniquely represented as:

$$Q\hat{R}S(t_z) = \sum_{n=0}^{M-1} C_n H(t_z) \quad (3)$$

with $M$ being equal to the discrete signal length. The Hermite coefficients are denoted with $C_n$. It is crucial to use the points $t_z$ in the calculation of Hermite basis functions. Note that the signal should be resampled at these points, if it is already uniformly sampled [12]. The following formula is used to calculate $C_n$ and it is based on the Gauss-Hermite quadrature method [11]-[19]:

$$C_n = \frac{1}{\sqrt{2^n n! \sqrt{\pi}}} \sum_{z=1}^{S} \frac{2^{S-1} S! \sqrt{\pi}}{S^2 P_{S-1}^2(t_z)} (QRS(t_z) e^{\frac{t_z^2}{2}}) P_n(t_z) \quad (4)$$

$S$ is the number of samples of the signal $QRS(t)$. It can be shown that under previous assumptions regarding the sampling points, $S = N$ holds. By using equation (2) and the following expression:

$$\xi_{N-1}^n(t_z) = \frac{H_n(t_z)}{H_{N-1}^2(t_z)} \quad (5)$$

the simplified form of expression (4) is obtained:

$$C_n \approx \frac{1}{N} \sum_{z=1}^{N} \xi_{N-1}^n(t_z) QRS(t_z) \quad (6)$$

### III. THE SVM CLASSIFICATION BASED ON HERMITE TRANSFORM FEATURES

Consider the data sets $(d_1, c_1), (d_2, c_2), \ldots, (d_n, c_n)$ where $d_n$ represents the input data and $c_n$ are the class labels which can take two values: 1 or -1. The aim of an automatic classificator is to design a decision algorithm which sorts the input data points in the class that they belong to. In order to design a good classificator, the maximum-margin hyperplane has to be introduced. If the training data is linearly separable, two parallel hyperplanes can be selected so that the data points of each class meet the following set of expressions [4], [5], [21]:

$$\begin{aligned} w^T d + b \geq 1, \text{ if } c_n = 1 \\ w^T d + b \leq 1, \text{ if } c_n = -1 \end{aligned} \quad (7)$$

$w$ is the weight vector and $b$ is the bias. The maximum-margin hyperplane is the one that lies halfway between the two selected hyperplanes [4], [5]. The distance between the data points that belong to different data classes is denoted with $l$ and expressed using equation (8):

$$l = \frac{2}{\|w\|} \quad (8)$$

By maximizing the margin, the possibility to get classification errors is decreased. A maximal $l$ is obtained when the denominator is minimal, respectively [4]:

$$\min \frac{1}{2} \|w\|^2 \quad (9)$$

where the requirement $c_n(w^T d_n + b) \geq 1$ needs to be met. For the purpose of minimization, the Lagrange multiplier $\eta_n$ is introduced [19]. Now the optimization problem becomes as follows:

$$\max \sum \eta_n - \frac{1}{2} \sum \sum \eta_n \eta_m c_n c_m d_n^T d_m \quad (10)$$

Knowing that $\eta_n \geq 0$ and $\sum \eta_n c_n = 0$, by analyzing equation (10) the following expressions are obtained:

$$\begin{aligned} w = \sum \eta_n c_n d_n \\ b = c_{sv} - w^T d_{sv}, \forall d_{sv}, \eta_{sv} \neq 0 \end{aligned} \quad (11)$$

The values of $d_{sv}$ that meet $\eta_{sv} \neq 0$ are the support vectors. The support vectors are the closest data points to the separating hyperplane and represent the main component of the decision function $f(d)$ [20]:

$$f(d) = \sum \eta_n c_n d_n^T d + b \quad (12)$$

Using this function, the class label for each data point $d$ can be determined. Every linearly separable data will be successfully separated by applying equation (12). But, each classification solution can further be improved to avoid ambiguity and misclassification due to the influence of noise or for the cases of data that cannot be separated successfully using the proposed solution. Therefore, two new variables are introduced: the regularization parameter $P$ and $\delta$. The optimization problem can be rewritten in the form [21]:

$$\frac{1}{2} w^T w + P \sum \delta_n, \text{ where } c_n(w^T d_n + b) \geq 1 - \delta_n, \delta_n \geq 0 \quad (13)$$

The Langrange multiplier introduced in equation (10) in this case can take values in the range $[0, P]$. The input training data becomes separable when translated to a higher dimensional space by defining $d_n^T d_m$ in terms of Kernel function $K(d_n, d_m)$.

If we represent the signal of interest using equation (3) which is obtained when the signal is expanded in the Hermite transform basis, then the coefficients $C_n$ are the features used in the classification.

## IV. EXPERIMENTAL RESULTS

Two sets of data made of five signals each, are analyzed in this paper. One of the data sets contains the QRS complexes of healthy people represented in the Hermite transform domain, while the other data set is the Hermite transform representation of the irregular QRS complexes showing heart health anomalies.

The performance of the proposed method for all the available data is shown on Figure 1. All the diseased QRS complexes are detected and classified correctly, including the most critic QRS complex which is zoomed in Figure 1. The diseased QRS complexes are marked in red, while the healthy ones are plotted in green. The support vectors are represented using circles. The accuracy of detection of irregular QRS complexes is 100%. However, if observed as a general classificator between healthy and irregular QRS complexes, then we might introduce the following terms: true positive (TP), true negative (TN), false positive (FP) and false negative (FN). The total number of QRS complexes is 161. The number of healthy QRS complexes is 72, out of which 35 are classified as healthy (TN) and the rest of them is misclassified (FP). All the irregular (diseased) QRS complexes are classified as diseased, which makes the number of correctly classified diseased QRS complexes (TP) 89. There are 0 false negatives (FN). As the aim of this paper is the detection of irregular i.e. diseased QRS complexes, any detection of this kind is referred to as positive. Now that all this information is obtained the efficiency of the proposed method for classification between healthy/regular and diseased/irregular complexes can be calculated as follows:

$$\frac{TP+TN}{total} \cdot 100\% = \frac{89+35}{161} \cdot 100\% = 77\% \quad (14)$$

The true positive rate can be expressed as follows:

$$TP/total\_diseased = 1 \quad (15)$$

while the rate of false positives is:

$$FP/total\_healthy = 1.06 \quad (16)$$

This false positive rate amount decreases the efficiency of the method for classification. The main factor that influences the rate of misclassification is the presence of noise: motion artifacts, baseline wander, electrode contacts, measurement equipment noise, etc. Signals should be acquired in specialized laboratories in order to test the performance of any classificator.

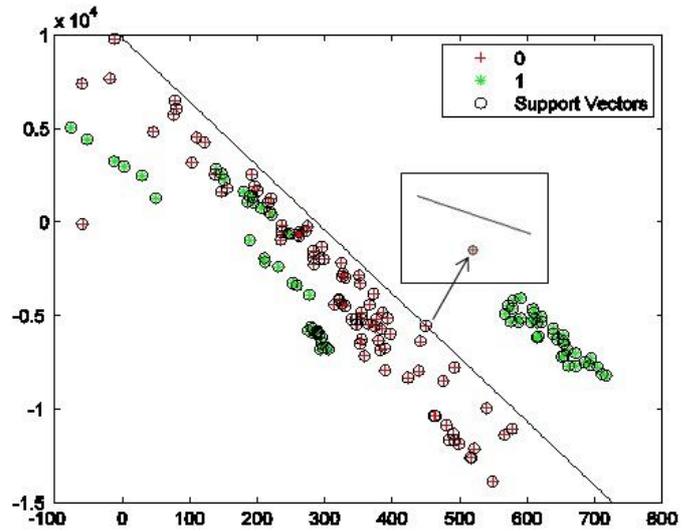

Figure 1. The detection of diseased QRS complexes

However, if we take into account these facts, the proposed classificator can give satisfactory results. For some cases, it can classify the two data sets with 100% of accuracy as shown on Figure 2. Two signals from both classes are selected and analyzed. The pair of signals is processed and the steps are displayed within the figure. The time domains of signals are shown on the top sub-figures of the first and the last figures, while the bottom sub-figures represent the Hermite transform coefficients of all the detected QRS complexes belonging to the appropriate signals. Observe that the Hermite coefficients of QRS complexes belonging to healthy (left figure) and diseased (right figure) people, differ significantly which indicates their potential for feature generation. The second figure shows the output of the classificator for the observed signals. The presented case is classified correctly.

## V. CONCLUSION

In this paper, the detection of irregular (diseased) QRS complexes using SVM is proposed. Two data sets were analyzed. One data set contains QRS complexes of healthy persons, while the other QRS complexes are irregular showing certain anomalies. To ensure fast data analysis and transmission, a reduced number of features is employed by representing the data in the Hermite transform domain. The performance of the proposed method is evaluated real-world signals. The method proved to be very efficient when used for detection of anomalies. The accuracy in this case is 100%. Future work could be oriented to decrease the number of misclassified healthy QRS complexes, which can be achieved by including, additional parts of ECG signals in the analysis and classification.

ACKNOWLEDGMENT

This work is supported by the Montenegrin Ministry of Science, project grant funded by the World Bank loan: CS-ICT "New ICT Compressive sensing based trends applied to: multimedia, biomedicine and communications".

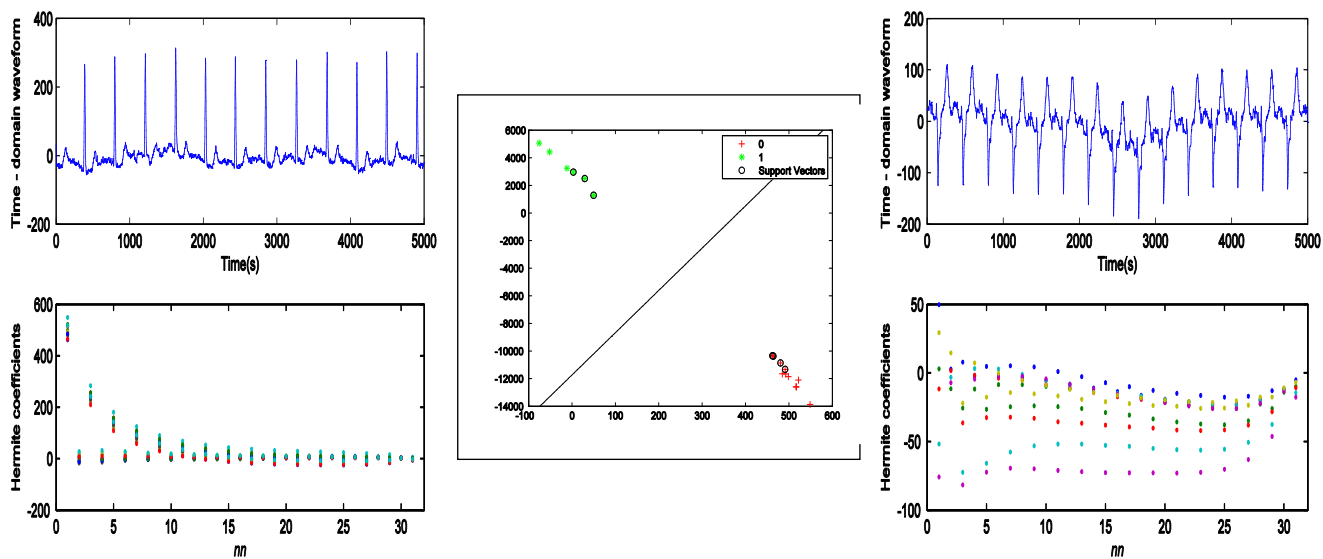

Figure 2. The classification healthy – diseased (first figure – Time domain waveform of healthy people ECG signal (top sub-figure) and Hermite coefficients of its QRS complexes (buttom sub-figure)), second figure – The performance of the classificatory, third figure – Time domain waveform of diseased people ECG signal (top sub-figure) and Hermite coefficients of its QRS complexes (buttom sub-figure))).